%% file: yumi_hri2018_header.tex
\documentclass[sigconf]{acmart}

\renewcommand\footnotetextcopyrightpermission[1]{} 
\usepackage{booktabs} 
\usepackage{tikz}
\usepackage{amsmath}
\usepackage{algorithm}
\usepackage[noend]{algpseudocode}
\usepackage{lipsum,graphicx}
\usepackage{subcaption}

\usepackage{xcolor}
\usepackage{pgfplots}

\makeatletter
\def\BState{\State\hskip-\ALG@thistlm}
\makeatother

\algnewcommand{\algorithmicand}{\textbf{ and }}
\algnewcommand{\algorithmicor}{\textbf{ or }}
\algnewcommand{\OR}{\algorithmicor}
\algnewcommand{\AND}{\algorithmicand}
\algnewcommand{\var}{\texttt}

\algnewcommand\algorithmicforeach{\textbf{for each}}
\algdef{S}[FOR]{ForEach}[1]{\algorithmicforeach\ #1\ \algorithmicdo}

\newcommand{\inote}[1]{$\ll$\textcolor{cyan}{Iolanda}$\gg$}
\newcommand{\enote}[1]{$\ll$\textcolor{red}{Elena}$\gg$}
\newcommand{\dnote}[1]{$\ll$\textcolor{magenta}{Dimos}$\gg$}
\newcommand{\onote}[1]{$\ll$\textcolor{magenta}{Olov}$\gg$}
\newcommand{\hnote}[1]{$\ll$\textcolor{magenta}{Hakan}$\gg$}

\makeatletter
\def\@copyrightspace{\relax}
\makeatother
\setcopyright{none}

\makeatletter
\def\runningfoot{\def\@runningfoot{}}
\def\firstfoot{\def\@firstfoot{}}
\makeatother 

\makeatletter
\patchcmd{\maketitle}{\@copyrightspace}{}{}{}
\makeatother

\makeatletter
\renewcommand\@formatdoi[1]{\ignorespaces}
\makeatother

\begin{document}

\title{A Comparison of Visualisation Methods for Disambiguating Verbal Requests in Human-Robot Interaction}


\author{Elena Sibirtseva}
\affiliation{%
   \institution{KTH Royal Institute of Technology}
   \city{Stockholm} 
   \country{Sweden} 
   \postcode{SE-10044}
 }
 
\author{Dimosthenis Kontogiorgos}
\affiliation{%
   \institution{KTH Royal Institute of Technology}
   \city{Stockholm} 
   \country{Sweden} 
   \postcode{SE-10044}
 }
 
 \author{Olov Nykvist}
\affiliation{%
   \institution{KTH Royal Institute of Technology}
   \city{Stockholm} 
   \country{Sweden} 
   \postcode{SE-10044}
 }
 
\author{Hakan Karaoguz}
\affiliation{%
   \institution{KTH Royal Institute of Technology}
   \city{Stockholm} 
   \country{Sweden} 
   \postcode{SE-10044}
 }
 
 \author{Iolanda Leite}
\affiliation{%
   \institution{KTH Royal Institute of Technology}
   \city{Stockholm} 
   \country{Sweden} 
   \postcode{SE-10044}
 }
 
\author{Joakim Gustafson}
\affiliation{%
   \institution{KTH Royal Institute of Technology}
   \city{Stockholm} 
   \country{Sweden} 
   \postcode{SE-10044}
 }
\author{Danica Kragic}
\affiliation{%
   \institution{KTH Royal Institute of Technology}
   \city{Stockholm} 
   \country{Sweden} 
   \postcode{SE-10044}
 }

\begin{abstract}

Picking up objects requested by a human user is a common task in human-robot interaction. When multiple objects match the user's verbal description, the robot needs to clarify which object the user is referring to before executing the action. Previous research has focused on perceiving user's multimodal behaviour to complement verbal commands or minimising the number of follow up questions to reduce task time. In this paper, we propose a system for reference disambiguation based on visualisation and compare three methods to disambiguate natural language instructions. In a controlled experiment with a YuMi robot, we investigated real-time augmentations of the workspace in three conditions -- mixed reality, augmented reality, and a monitor as the baseline -- using objective measures such as time and accuracy, and subjective measures like engagement, immersion, and display interference. Significant differences were found in accuracy and engagement between the conditions, but no differences were found in task time. Despite the higher error rates in the mixed reality condition, participants found that modality more engaging than the other two, but overall showed preference for the augmented reality condition over the monitor and mixed reality conditions.
\end{abstract}


%
%



\maketitle

\input{yumi_hri2018_BODY}

\bibliographystyle{ACM-Reference-Format}
\bibliography{bibliography} 

\end{document}

%% file: yumi_hri2018_BODY.tex
\section{Introduction}

Picking up objects is a common task for robots that work alongside people in home and workplace environments. A typical human-robot interaction task consists of a robot assisting a worker as a third hand, retrieving requested items out of a variety of similar objects. 

It is intuitive for humans to use natural language when their hands are busy and they cannot point at the target object. However, such interactions can often lead to ambiguous requests because of speech recognition and language understanding errors, limitations in the robot's understanding of the scene or the presence of similar objects in the workspace. 

Previous research has tackled the problem of disambiguating requests from two different perspectives. One perspective aims to reduce ambiguity by asking follow-up  questions. However, the more clarification questions the robot asks, the longer the task takes and the risk of speech recognition errors is likely to increase. Previous work that focuses on minimizing the number of follow-up questions has shown that verbal interactions increase task time and can influence accuracy \cite{whitney2017}. An alternative approach consists of employing visualisation techniques such as augmented \cite{andersen2016} or mixed reality \cite{frank2016} to augment the scene with the robot's or human's intentions. 
While the first few works in this direction have started to appear \cite{chadalavada2015, rosen2017, pereira2017}, the effects of augmenting the workspace to disambiguate user verbal requests are still unknown. 

\begin{figure}[t!]
  \includegraphics[width=0.47\textwidth]{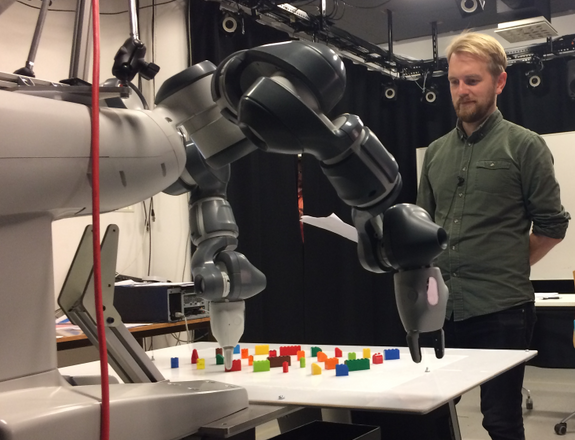}
  \caption{A participant interacting with the YuMi robot in our experiment using verbal requests to exchange Lego blocks.}
  \label{fig:erik}
\end{figure}

In this paper, we performed an experiment to investigate different real-time visualisation modalities for disambiguating verbal requests in an object retrieval task. We developed a system that, in the presence of ambiguous verbal requests, highlights candidate objects and updates this selection as the user refines the target object description with new verbal requests. Using this system, we tested three modalities for providing visual information about the candidate objects that the robot is considering in the workspace: augmented reality (using a projector), mixed reality (using Microsoft HoloLens), and a side monitor as the baseline condition.


Our experimental setup consisted of an ABB YuMi robot and a table with Lego blocks (Figure \ref{fig:erik}). The robot and the human took turns while requesting Lego blocks to pick up. Participants had to verbally explain which Lego block they wanted, using shape and colour information, and were able to perceive by looking at the real-time visualisation the robot's hypothesis about which objects match that description. We intentionally designed this setup to include blocks that would originate ambiguous requests.

In a within-subjects experiment, we collected task times and accuracies, as well as subjective metrics such as engagement, task observability, display interference and personal preferences. The results of the study showed no significant difference in task time between three conditions. Furthermore, accuracy significantly decreased in the mixed reality condition; however, participants regarded this condition as the most engaging compared to the other two. As anticipated, the augmented reality condition provided better observability of robot's behaviour and was considered less disruptive. Finally, the augmented reality interface was preferred by most participants and viewed as the most natural and easy to understand visualisation method.


\section{Related work}
In object retrieval tasks natural language is commonly interpreted into semantically informed representations of the physical space between humans. In human communication, language grounding refers to establishing a ``common ground'' and understanding that both parts refer to the same object or concept \cite{clark1996using}. There have been early attempts in linguistics research in the '70s \cite{winograd1972understanding}, where users interact with a machine that can understand simple references to objects. Further attempts were made to solve the problem using multimodal features \cite{bolt1980put}, and disambiguate verbal references to objects in a virtual space.

Humans use various methods to establish common ground when they instruct each other in collaborative object retrieval tasks. Common problems occur when object ambiguity is encountered. This makes it more challenging to establish grounding. Li et al. \cite{li2016spatial} experimented with natural language instructions to investigate the effect of object descriptors, perspective and spatial references and found that ambiguous sentences take more time to process.

Establishing language grounding, particularly in situated human-robot dialogue, can be challenging. Robots need to perceive human behaviour and build internal representations and spatial semantic understanding based on human intentions \cite{steels2012language}. Recent research has approached the problem linguistically and through incremental reference resolution \cite{chai2014collaborative,skantze2010jindigo,kennington2017simple}, spatial references \cite{paulgrounding,guadarrama2013grounding}, modelling uncertainty \cite{hough2017s}, but also through past visual observations \cite{ijcai2017-629}.

Other approaches have considered multimodal features to disambiguate verbal references to the physical space. Several studies have investigated methods such as eye gaze and pointing gestures to disambiguate referring expressions to objects in the shared space between humans and robots \cite{mehlmann2014exploring,renner2014spatial,sauppe2014robot,admoni2016modeling}, and explored non-verbal communicative behaviours to achieve grounding.

Whitney et al. \cite{whitney2017} used language and pointing gestures at specific objects when there was ambiguity in the human request. A POMDP based framework was developed in order to balance out the trade off between gaining additional information and the risk of facing speech-to-text failures. However, such an interaction can take a lot of time and would be infeasible with a larger amount of objects. One of the ways to solve this is to visualise the current state of the robot's understanding of the request.

Several works have shown effectiveness of using projector based approaches to augment robot's intentions into the shared workspace \cite{Ruffaldi2016, pereira2017, andersen2016, chadalavada2015}. In particular, Andersen et al. \cite{andersen2016} proposed an object-aware projection technique which takes into account the 3D nature of the environment. As a possible use-case they proposed a car assembly line, where car doors are transported on a conveyor belt and both human and robot have to engage as co-workers on the door. Augmented reality is used to mark the parts that the robot is currently working on. A user study was performed, in which the task was to either rotate or move a white box, based on the instructions provided by one of the three interfaces: projector, monitor display, and text description. The evaluation of this study showed that the augmented reality approach scored higher in user effectiveness and user satisfaction compared to a baseline condition.

\begin{figure}[t!]
  \includegraphics[width=0.47\textwidth]{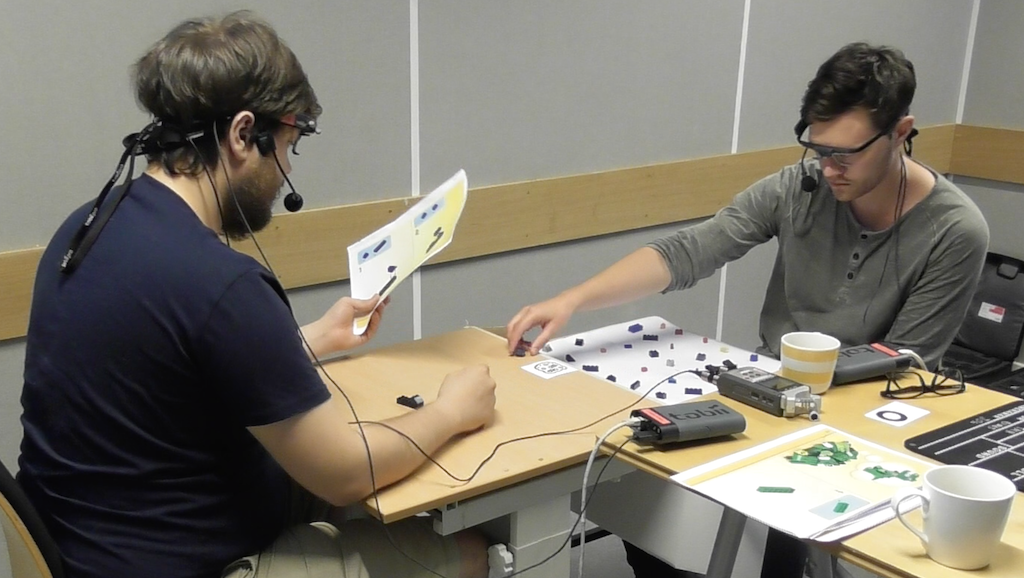}
  \caption{Human-human interaction pilot study to investigate the most common verbal references used by participants in the task.}
  \label{fig:humanhuman}
\end{figure}

Moreover, another successful application of augmented reality for showing robot's intent was demonstrated in \cite{chadalavada2015}, where Chadalavada et al. equipped a robotic fork-lift with a projector to visualise its future trajectory a few meters ahead. The results of the human study showed that by visualising the robot's intent, they achieved significant increase in predictability and transparency; the attributes most crucial for the acceptance of the robots in the workspace. 


%
The application of mixed reality to human-robot interaction is an emerging field of research and shows promising results. For instance, Rosen et al. \cite{rosen2017} proposed a mixed reality framework to visualise future trajectories of the robot motion. To evaluate the performance of the proposed framework, they conducted a study where participants were asked to detect collisions of robot arm motions using three interfaces: no visualisation, monitor 3D point cloud view from a Kinect sensor, and mixed reality with Hololens. The authors found that the mixed-reality condition for this specific task is faster, more accurate, and subjectively more enjoyable.

\section{Human-human pilot study}
In order to inform the design of our reference disambiguation visualisation system, we first carried out a human-human interaction pilot study on a collaborative task involving object retrieval. We recruited 10 participants (5 pairs) that took turns in asking for and fetching Lego blocks of various colours and shapes to build a model. Since we were interested in verbal references, we asked participants to avoid pointing and instead use only verbal instructions (see Figure \ref{fig:humanhuman}).

We found that most participants used the terms colour and shape to describe the blocks, which informed the design of the system described in the next section. 
Using an off-the-shelf speech recognition system to transcribe the collected audio data resulted in many incorrect object descriptors, possibly augmented by the fact that none of the participants were native English speakers. We therefore decided to make the language understanding module of our system controlled by a wizard. 

\section{System description}

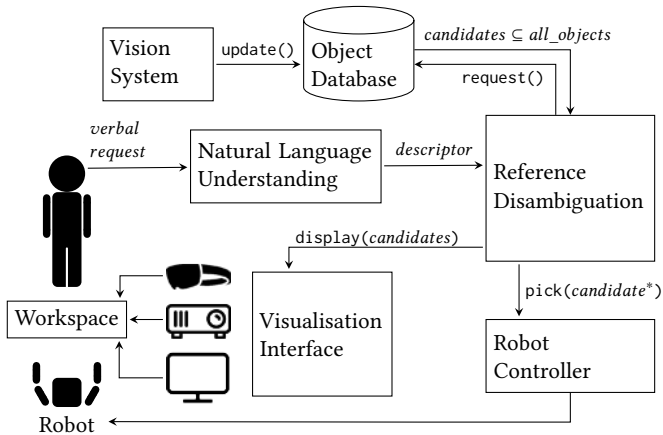
\begin{figure}[t!]
\centering
\input{system.tex}
\caption{The architecture of the proposed system for visualising ambiguous fetching requests.}\label{fig:system}
\end{figure}

We designed and implemented a system that takes user verbal requests as an input and processes them through the modules depicted in Figure \ref{fig:system}. If there is ambiguity in the request (i.e., more than one object matches the colour or shape described by the user) the system highlights the candidate objects using the visualisation interface while awaiting for further verbal commands that refine the request. This process continues until there is no more ambiguity and the robot is able to pick up the target object. 

When the user makes a verbal request explaining which block the robot should fetch, a human wizard performs the \textbf{natural language understanding} to extract colour and shape object descriptors supported by the system. This module is the only wizarded component of the system.


Given the object descriptors, the \textbf{reference disambiguation} module queries the object database to get the candidate objects that match the provided descriptors. The \textbf{object database} stores colour and shape attributes, 3D positions and rotation with respect to the robot of all the objects present in the workspace. The object attributes are continuously updated by the \textbf{vision system}, which uses a Microsoft Kinect sensor. The vision system works as follows. A region-of-interest (ROI) that represents the robot's workspace is defined on the image with the objects. Individual objects within the workspace are continuously segmented using colour segmentation and morphological operations. Finally, 3D position and rotation estimates of the Lego blocks with regard to the robot are calculated using the depth information from the Kinect sensor.

After receiving the candidate objects from the database with updated positions, the reference disambiguation module resolves object references to highlight the relevant objects using one of the visualisation interfaces. If the object descriptors cannot further disambiguate the available objects (i.e. when more than one of the same colour or shape exist), then numbers are displayed next to the objects. When there exists only one available object fitting the descriptor, the pick up command and the object coordinates are sent to the robot controller. Alg. \ref{resolve} summarizes this procedure.

The \textbf{robot controller} module is responsible for receiving the object coordinates from the reference disambiguation module and performing motion planning to pick up the target object and place it on a side bin. In our implementation with the YuMi robot, the low-level arm motions are planned and executed using an open source ROS-based motion planner \cite{chitta2012moveit}. The corresponding arm for the action is selected based on the target's location. 

Finally, one of the \textbf{visualisation interfaces} highlights the candidate objects that, in the robot's perceptive, match the human's verbal request of one or several objects. We developed interfaces that support three different visualisation methods: a side monitor, augmented reality, and mixed reality. These methods will be described in more detail in the next section.

\begin{algorithm}[t!]
\caption{Visualisation of object highlighting extracted by object descriptors from human instructions.}\label{resolve}
\begin{algorithmic}[1]
\State candidates $\gets$ []
\Procedure{VISUALISE(objectDescriptors)}{}
	\State candidates $\gets$ queryObjectDB(objectDescriptors)
    \State display(candidates)
    \If{len(candidates) = 1}
    	\State pick(candidate*)
        \State updateObjectDB()
    \EndIf
    \If {singleShape(candidates)\\
        \hspace{0.9cm}\AND singleColour(candidates)}
        	\State displayIDs(candidates)
        \EndIf
\EndProcedure
\end{algorithmic}
\end{algorithm}

\begin{figure*}[t]
  \centering
  \subcaptionbox{Monitor (MO)\label{cond:MO}}			[.3\linewidth][c]{\includegraphics[width=.3\linewidth]{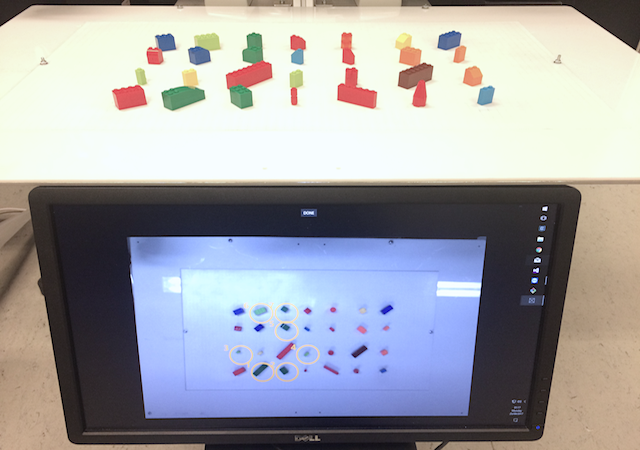}}\quad
  \subcaptionbox{Augmented Reality (AR)\label{cond:AR}}[.3\linewidth][c]{\includegraphics[width=.3\linewidth]{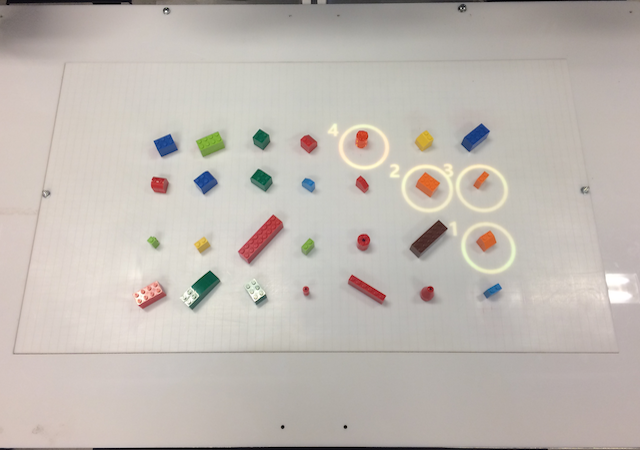}}\quad
  \subcaptionbox{Mixed Reality (MR)\label{cond:MR}}		[.3\linewidth][c]{\includegraphics[width=.3\linewidth]{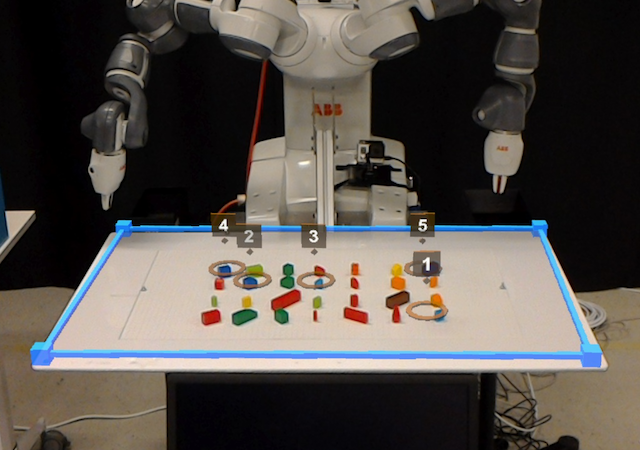}}\quad
  \caption{The three visualisation methods evaluated in our experiment.}
  \label{fig:conditions}
\end{figure*}

\section{Evaluation}
Our evaluation scenario consisted of a YuMi robot capable of retrieving Lego blocks following participants' verbal instructions. Using this scenario, we evaluated the reference disambiguation system described in the previous section by comparing three different visualisation modalities using a within-subjects design:

\begin{itemize}
\item \textbf{{Monitor (MO)}}. A monitor near the workspace streaming the video from a web-camera directed at the table from the top (Figure \ref{cond:MO}). The candidate object highlights were overlayed on the video stream. The monitor was placed in a best possible position we encountered without interfering with the robot's manipulations of the objects.

The monitor was positioned in a place where the cognitive mapping of the physical objects to the monitor is realised in an optimal way considering the available positions in the setup.

\item \textbf{Augmented Reality (AR)}. In this condition, we used a projector which augmented the candidate highlights directly on the physical workspace (Figure \ref{cond:AR}).

\item \textbf{Mixed Reality (MR)} we used a commercial head-mounted display\footnote{https://www.microsoft.com/en-us/hololens} to show the candidate objects by merging the virtual 3D highlights into the real world (Figure \ref{cond:MR}). The virtual workspace was initially calibrated to align with the real workspace using a fiducial marker, but the continuous tracking was performed based on the spatial mapping provided by the mixed reality device.
\end{itemize}



\subsection{Hypotheses}
We formulated the following hypotheses for this experiment:
\begin{itemize}
\item \textbf{H1:} Participants will take longer to complete trials in the MO condition than in the AR and MR conditions.
\item \textbf{H2:} Participants will commit fewer mistakes in the AR and MR conditions than in the MO condition.
\item \textbf{H3:} Participants will consider the MR condition more engaging than the MO and AR conditions.
\item \textbf{H4:} Participants will consider the AR condition less disruptive compared to the other two conditions.
\item \textbf{H5:} Participants will prefer the AR and MR conditions to the MO condition.
\end{itemize}

We base \textbf{H1} and \textbf{H2} on the premise that if participants need to perform spatial mapping from the shared workspace to the the monitor, this will potentially contribute to a higher cognitive load. Similarly, because participants need to look away from the workspace and back at the monitor in MO, this will likely increase the number of errors.
Our reasoning for establishing \textbf{H3} and \textbf{H5} is drawn from previous research showing that mixed reality applications can improve user experience \cite{rosen2017,pereira2017}.
\textbf{H4} is argued for by reasoning that the augmented reality condition will enable participants to dedicate full attention to the workspace.

\subsection{Participants}
A total of 29 subjects (12 female, 17 male), with ages between 22 and 50 ($M=28.8$), were recruited for this experiment using mailing lists and flyers. To be able to participate in the experiment, subjects needed to be fluent in English, not have any colour vision deficiency and not wear glasses (due to difficulties wearing the mixed reality device).

On a scale from 1 to 5 (with 1 representing high), participants' familiarity with digital technology was 1,8. Additionally, 21 out of the 29 participants had tried Augmented or Virtual Reality before, while 9 out of 29 had interacted with a robot before.


\subsection{Procedure}
Upon arrival, participants were given a consent form and instructions about the experimental process. They were instructed to ask a robot to pick up a set of Lego blocks using only colour and shape descriptors without pointing or using spatial references (e.g. ``the block next to the red one'').

After that, participants went through a training phase with the experimenter where they picked up Lego blocks in turns as if they were interacting with the robot to get familiar with the task. Before each experimental trial, participants were given a piece of paper listing images of the blocks they would have to request from YuMi. Each trial consisted of 15 turns where the human participant and the robot took turns while requesting Lego blocks from each other from the shared workspace. The participant started first and requested in each trial 8 objects and the robot 7. While participants had to make their requests using verbal descriptors, YuMi's requests simply consisted of highlighting the target block using the active visualisation modality in that trial. This type of request was simply included in the experiment to ensure that participants took actions in the physical workspace and avoid, for example, that in the MO condition they simply followed the video feed shown in the monitor. Each trial took 8 minutes on average. Participants filled task questionnaires after each trial and a final questionnaire at the end of the experiment.





We used a balanced Latin square array to counterbalance the order of conditions being tested by each participant and avoid order effects. The initial arrangement of Lego blocks on the table was randomised in each session, meaning that participants did not use the same arrangement twice. An experimenter was always present in the room to ensure blocks were removed from the table in cases of occasional grasping errors and intervene if necessary. 	

We recorded audio and video in all sessions and logged time measurements and object requests for further analysis.

\subsection{Measurements}
To investigate the presented hypotheses, we collected both objective and subjective measures. From the interaction logs, we extracted the average \textbf{request time} per object considering the portions of the task where the participant describes a Lego block for the robot to pick up to the moment the Reference Disambiguation module sends a pick request to the robot controller (note that this excludes the robot's action completion time). The first two human block requests were excluded from each trial because their duration might have been biased by the fact that participants were still getting used to the modality/device (especially in the MR condition).
A human annotator analysed the video recordings and counted the number of incorrect task executions per trial, i.e. when participants either described the wrong block to the robot or picked a block different than the requested one. This frequency was normalised by the total number of turns of each trial and will be referred to as the \textbf{error rate} per trial.


After participating in each trial, participants answered subjective questions extracted from The Presence Inventory \cite{lombard2009} and the Presence Questionnaire \cite{witmer1998} about their perceived \textbf{engagement}, \textbf{observability} (i.e. how well they could observe the robot's behaviour) and \textbf{display interference} (i.e. the degree to which the visual display quality interfered with or distracted from task performance). Participants answered these questions using a 7-point Likert scale where 1 meant ``Not at all'' and 7 meant ``Very much''. At the end of the experiment, they answered additional questions regarding their \textbf{preferences} such as which condition they preferred, which condition they found easiest to perform the task and which condition would they pick to work with in the future. The final survey also included open ended questions about the advantages and disadvantages of each modality, as well as generic questions about participants' previous experience with robots, video games and AR/VR devices.


\section{Results}
This section presents the results of the objective and subjective measures collected in the experiment.


\subsection{Objective Measures}
The objective measures were analysed using one-way repeated measures ANOVA. Because the first trial of each participant took longer than the other two trials in general, for analysing the request time variable, we included the order of the trials as a within-subjects factor. Post hoc tests were performed using the Bonferroni correction. Figure \ref{fig:chart_obj} complements the results presented below.

\subsubsection{Request Time} For the portions of the task where participants described a Lego block for the robot to pick, we found no significant main effect of condition, $F(2, 14)= 1.19, p= .33 , \eta^{2}= .15 $. A significant order effect was found, $F(2, 14)= 11.43, p < .05, \eta^{2}= .62 $, such that the average duration of request turns was higher in the first trial ($M=$ 20.61, $SE=$ 1.48) than in the second ($M=$ 16.84, $SE=$ .53) and third ($M=$ 14.36, $SE=$ .44) trials, regardless of condition. Post hoc tests revealed no significant differences between the first and second trials ($p = .18$), but a significant difference between the second and third trials ($p <$ .05), as well between the first and third trials ($p <$ .05). No significant interaction effect was found between condition and trial, $F(4, 28)= .57, p = .69, \eta^{2}= .08$.


\subsubsection{Error Rates} We found a significant effect of condition, $F(2, 56)= 3.22, p < .05, \eta^{2}= .10 $, such that in the AR condition the participants had the lowest error rates ($M=$ .01, $SE=$.01), followed by the MO condition ($M=$ .02, $SE=$.01) and then the MR condition ($M=$ .04, $SE=$.01). Post hoc tests revealed that the AR condition had significantly lower error rates than the MR condition ($p = 1.0$), but no significant differences were found between error rates between MO and MR, nor MO and AR. 

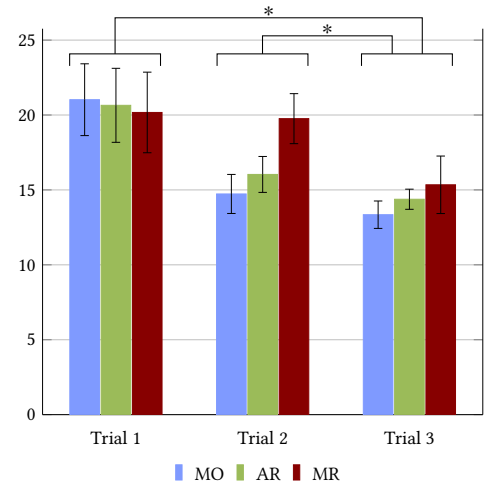
\begin{figure}[t!]
  \input{chart_obj.tex}
  \caption{Average duration (in seconds) of participants' verbal request by trial number and condition. (*) denotes p < .05.}
  \label{fig:chart_obj}
\end{figure}

\begin{figure}[t!]
  \input{chart_subj.tex}
  \caption{Questionnaire responses for perceived Engagement, Observability and Display Interference. Ratings were provided on a 7-point Likert scale. (*) denotes p < .05}
  \label{fig:chart_subj}
\end{figure}
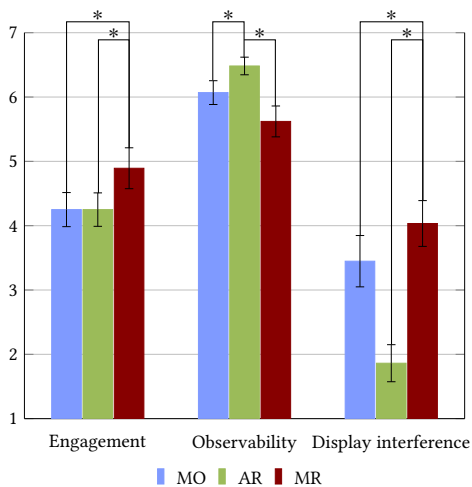

\subsection{Subjective Measures}
The subjective measures collected after each trial (engagement, observability and display interference) were analysed using one-way repeated measures ANOVA, and the multiple choice questions of the final survey we analysed using Chi Squared tests. When post hoc comparisons were done, we used the Bonferroni correction. The results reported here are summarised in Figure \ref{fig:chart_subj}.


\subsubsection{Engagement} We found a statistically significant effect of condition, $F(2, 54)= 4.93, p < .05, \eta^{2}= .15$, such that participants found the MR condition to be the more engaging ($M=$ 4.89 , $SE=$ .32) than both MO ($M=$ 4.25, $SE=$ .27) and AR ($M=$ 4.25, $SE=$ .26). Post hoc tests revealed that engagement ratings were significantly higher in the MR condition than both the MO and AR conditions ($p < .05$ in both comparisons), but no significant differences were found in perceived engagement between the MO and AR conditions ($p = 1.0$).

\subsubsection{Observability} There was a statistically significant effect of condition, $F(2, 56)= 8.74, p < .01, \eta^{2}= .24$, such that participants considered that they were best able to observe the robot's behaviour in the AR condition ($M=$ 6.48, $SE=$ .14), followed by the MO condition ($M=$ 6.07, $SE=$ .19) and finally the MR condition ($M=$ 5.62, $SE=$ .24). Post hoc tests showed that the AR condition was considered better to observe the robot's behaviour compared to the MO and MR conditions ($p < $ .05 for both comparisons), but no significant differences were found between the MO and MR conditions ($p =$ .19)

\subsubsection{Display interference} A statistically significant effect was found of condition, $F(2, 56)= 14.11, p < .001, \eta^{2}= .34 $. The condition in which the display less interfered with the task was the AR ($M=$ 1.86, $SE=$ .29), followed by the MO ($M=$ 3.45, $SE=$ .40) and then the MR ($M=$ 4.03, $SE=$ .36). Post hoc comparisons revealed that these differences were statistically significant between MO and AR ($p <$ .05), AR and MR ($p <$ .001), but not between MO and MR ($p =$ .64).

\subsubsection{Overall Preferences} There was a significant difference in the answers to ``In which condition did you prefer to use the robot?'' ($\chi^2 = 36.69, p < 0.001$) such that the highest number of participants preferred the AR condition. Similarly, in the responses to the question ``In which condition did you find it easiest to perform this task?'', participants found the AR condition significantly easier than the other two conditions ($\chi^2 = 18.29, p < 0.001$). Finally, we found a statistically significant difference in answers to the question ``Which condition would you pick to work with?'' ($\chi^2 = 27.79, p < 0.001$), such that the AR condition was the one participants would prefer to work with in the future.

\begin{table}
\centering
\caption{Preference Results (one participant did not answer one of the questions).}
\label{tab:questprefs}
\begin{tabular}{|c|c|c|c|} \hline
Question & MO & AR & MR\\ \hline 
Prefer &  1& 25& 3\\ 
Easiest &  4& 20 & 4\\ 
Use Again & 2 & 23 & 4\\ \hline
\end{tabular}
\end{table}

\section{Discussion}
Our first hypothesis stated that participants would take longer to complete the task in the MO condition compared to the AR and MR conditions. This hypothesis was not supported, as there were no significant differences between the request times between conditions. The significant difference between the average request duration in the first trial compared to the other two trials was likely caused by a learning curve on how to interact with the system: even though participants were told that YuMi was only capable of understanding shapes and colour descriptions, in the first trial participants tended to use other ways to describe the objects such as spacial references (e.g., ``the one closer to you'') that were not supported by the system. 

H2 stated that participants would commit fewer mistakes in the AR and MR conditions than in the MO condition, a hypothesis that was partially supported. Although the smallest error rates occurred in the AR condition, participants committed more task mistakes in the MR than in the MO condition. We believe that the errors in the MR condition were mainly a consequence of limitations of the mixed reality device such as limited field of view, which lead participants to sometimes lose their perspective of the entire workspace. Nevertheless, the average error rate was fairly low in all conditions. 

Despite the higher error rates in the MR condition, participants did find this condition more engaging than the other two conditions, a finding aligned with previous research on augmented reality in HRI \cite{pereira2017}. One of the mentioned advantages of the MR condition which might have contributed to higher engagement was the increased freedom to move around in the environment; regardless of their point of view, they were able to visualise the highlighted objects. However, it is also important to note that the higher engagement of this modality could have been caused by a novelty effect. Therefore, H3 (participants will consider the MR condition the most engaging) was supported.


In H4, we stated that the AR condition would be considered less disruptive than the other two conditions. This hypothesis was supported by our results for observability and display interference. Not surprisingly, in the open ended questions participants mentioned that because of the wearable device in the MR condition, and the fact that they had to switch their attention between the monitor and the workspace in the MO condition, these two conditions were more disruptive than the AR condition.

The questions regarding modality preferences followed the same trend as H4 and participants clearly chose the AR condition over the other two conditions. Many participants used words like ``natural', ``easy to understand'' and ``simple'' to characterize the AR condition. Some participants considered this modality to require the least cognitive load of all the conditions they interacted with. On the other hand, participants considered the MR condition to be more intrusive, with a limited field of view for the visualisation projection and somewhat uncomfortable to wear after some time because of its weight. While some of these disadvantages will become less evident with advances in hardware, mixed reality devices will likely remain more intrusive than the other two types of modalities we investigated. Regardless of these limitations, participants appreciated the ``portability'' aspect in the MR condition, especially when compared to the projector in the AR condition. The most common disadvantage identified in the MO condition was the need to map the scene back and forth between the monitor and the physical workspace. Participants who preferred the MO condition often did so for considering this modality to be the most familiar to them.

Our main goal was to investigate the impact of augmented and mixed reality visualisation methods when compared with typical ways of visualising information such as a monitor. As such, we deliberately decided not to include a control condition where the robot used pointing or follow up questions to disambiguate requests. Furthermore, it is important to note that without any sort of disambiguation requests, participants would not be able to complete parts of the task, since in each trial there was at least one situation where two objects had the same shape and colour.


\subsection{Limitations}
As one of the initial explorations in this domain, our experiment has several limitations that need to be addressed in future work. For example, we did not account for task difficulty (all the trials had similar levels of ambiguity), the objects were arranged in such a way that from most participants' viewpoints there were no occlusions, and the shared workspace consisted of a flat surface. As such, further research is needed to see whether the same results apply to more difficult tasks that would increase participants' cognitive load, as well as to more complex scenes where either because of the object placement or the nature of the projection surface, the 3D projections (only possible in the mixed reality condition) would play a more important role in the visualisations.

Finally, in the trial phase participants were able to practice the flow of the task with the experimenter, but we did not give them the opportunity to wear the mixed reality device until they actually had to use it in the trial. While most participants reported to have used other AR and VR devices before, the lack of experience with such interfaces might have an impact on participants' performance. In the attempt to account for this effect, we excluded the first two request turns of each trial, but a larger participant sample would have helped us to better understand whether previous experience with such devices influenced the results.



\subsection{Design implications}
Our findings indicate that the three investigated visualisation methods (monitor, augmented reality and mixed reality) are equally effective for displaying the robot's intentions in the presence of ambiguous requests. Nevertheless, other factors such as user experience, the nature of the task and practical considerations about cost and flexibility of the setup might affect the choice of one modality over another. This section discusses the advantages and disadvantages of each modality along these factors to inform future decisions of employing these methods in HRI scenarios.

\textbf{User experience.} While users found the mixed reality modality more engaging, not surprisingly they also considered it the most intrusive. Since engagement and attention are related concepts \cite{witmer1998}, mixed reality can be useful in tasks requiring the user to remain extremely focused. However, given the current hardware limitations in weight and field of view of these devices, mixed reality might not be suitable for very long tasks. As discussed in the limitations, the cognitive load in the monitor condition is likely to increase as task complexity increases, which might negatively affect users' engagement and task performance. As such, augmented or mixed reality modalities might be suitable for more complex tasks.



\textbf{Technical Considerations.} The mixed reality modality is better at dealing with occlusions and non-flat surfaces, but its limited field of view can become an issue in very large workspaces. These considerations are therefore relevant when considering the target application domain where the projections will be used. It is important to note, however, that with hardware improvements (which are likely to happen given the increasing research in this area) these considerations will tend to change over time.


\textbf{Practical Issues.} Although the monitor and the projector are more familiar and in general less expensive solutions, it should be noted that they are less flexible for requiring a permanent installation on top of the workspace. While this is not a problem for stationary workspaces, when considering, for example, fetching tasks with mobile robots, the lack of mobility in the setup can become an issue. In this case, a mixed reality solution becomes a clear choice.


\section{Conclusions}
In this paper, we investigated different visualisation methods for conveying to users which objects a robot is considering given verbal requests. We conducted a controlled experiment to compare three visualisation interfaces: mixed reality, augmented reality and a monitor as a control condition. Both objective (request time and error rate) and subjective measures (engagement, observability, display interference and preferences) were taken into account.

Our assumption was that mixed reality and augmented reality interfaces will decrease task time and increase accuracy compared to the control condition. However, the results of our findings showed no significant difference in task time related to condition. On the other hand, the mixed reality interface increased error rates compared to the other two conditions (although these were generally low). Despite this fact, participants found the mixed reality condition more engaging. Most participants preferred the augmented reality modality because they found it the easiest to use and less intrusive for this specific setting.

In future work, we will explore two different research directions. One of them is to explore benefits of the mixed reality in the tasks with irregular surfaces and object occlusion. Another relevant topic to investigate is the integration of other human perception modalities, such as pointing and gaze direction, to complement verbal requests and investigate the effects of visualisation methods for even more effective disambiguation.




%% file: system.tex
\usetikzlibrary{arrows,%
                petri,%
                topaths,
                calc,
                positioning,
                shapes.geometric}%
\begin{tikzpicture}[scale=0.97, transform shape,
>=stealth,
    node distance=3cm,
    database/.style={
      cylinder,
      shape border rotate=90,
      aspect=0.25,
      draw
    }]

\node[circle,fill,minimum size=5mm] (head) {};
\node[rounded corners=2pt,minimum height=1.3cm,minimum width=0.4cm,fill,below = 7pt] (body) {};
\draw[line width=1mm,round cap-round cap] ([shift={(2pt,-1pt)}]body.north east) --++(-90:6mm);
\draw[line width=1mm,round cap-round cap] ([shift={(-2pt,-1pt)}]body.north west)--++(-90:6mm);
\draw[thick,white,-round cap] (body.south) --++(90:5.5mm);

\node [draw, shape=rectangle] (ws) at (0, -2) {Workspace};

\node[rounded corners=2pt,minimum height=0.4cm,minimum width=0.4cm,fill, label=below:Robot] (ybody) at (0, -3) {};
\draw[line width=1mm,round cap-round cap, rotate=-30] ([shift={(7pt, 0pt)}]ybody.north east) --++(-90:3mm);
\draw[line width=1mm,round cap-round cap, rotate=30] ([shift={(-7pt, 0pt)}]ybody.north west)--++(-90:3mm);
\draw[line width=1mm,round cap-round cap] ([shift={(6pt, 6pt)}]ybody.north east) --++(-90:3mm);
\draw[line width=1mm,round cap-round cap] ([shift={(-6pt, 6pt)}]ybody.north west)--++(-90:3mm);

\node [draw, shape=rectangle, text width=2.4cm, minimum height=1cm] (nlu) at (3, 0.1) {Natural Language Understanding};

\node [draw, shape=rectangle, text width=2.1cm, minimum height=2cm] (rd) at (6.9, -0.2) {Reference\\Disambiguation};

\node[database,text width=1.3cm] (db) at (4, 1.5) {Object\\Database};

\node [draw, shape=rectangle, text width=1.3cm, minimum height=1cm] (vs) at (1.25, 1.5) {Vision\\ System};

\node [draw, shape=rectangle, text width=2.1cm, minimum height=1cm] (cs) at (6.9, -2.5) {Robot\\Controller};

\node [draw, shape=rectangle, text width=1.7cm, minimum height=1.7cm, align=left] (vi) at (3.5, -2.2) {Visualisation\\Interface};

\node[inner sep=0pt] (mo) at (1.8, -2.8) {\includegraphics[scale=0.5]{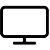}};
\node[inner sep=0pt] (ar) at (1.8, -2.0) {\includegraphics[scale=0.5]{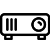}};
\node[inner sep=0pt] (mr) at (1.8, -1.4) {\includegraphics[scale=0.5]{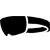}};

\draw [shorten >=1pt,shorten <=15pt,->] ([yshift=2pt]head.west) to node[above, scale=.8, text width=1cm] {$verbal$\\$request$} (nlu);

\draw [shorten >=1pt,shorten <=1pt,->] (nlu) to node[above, scale=.8, text width=1cm, xshift=-0.15cm] {$descriptor$} ([yshift=9pt]rd.west);


\draw [shorten >=1pt,shorten <=1pt,->] ([xshift=-20pt]rd.south) to node[right, scale=.8] {\texttt{pick($candidate^{*}$)}} ([xshift=-20pt]cs.north);

\draw [shorten >=1pt,shorten <=1pt,->] (mr) -| ([xshift=48pt]ws);
\draw [shorten >=1pt,shorten <=1pt,->] (mo.west) -| ([xshift=63pt]ws);
\draw [shorten >=1pt,shorten <=1pt,->] (ar) to (ws);

\draw [shorten >=1pt,shorten <=1pt,->] (vs) to node[above, scale=.8] {\texttt{update()}} (db);

\draw[shorten >=1pt,shorten <=1pt,->] ([yshift=-23pt]rd.west) -| node[above right, scale=.8, yshift=-0.1cm] {\texttt{display($candidates$)}} ([xshift=-13pt]vi.north);

\path[draw,<-] (rd.north) |- node[above, scale=.8, xshift=-0.9cm] {$candidates \subseteq all\_objects$} ([yshift=0.2cm]db.east);
\path[draw,<-] (db.east) -| node[below, scale=.8, xshift=-0.9cm] {\texttt{request()}} ([xshift=-0.2cm]rd.north);

\path[draw,->, shorten >=10pt] (cs.south) -- ++(0,-0.4cm) -- ++(-2.1cm,0) -- ([yshift=-0.4cm]ybody.east);

\end{tikzpicture}

%% file: chart_obj.tex
\usetikzlibrary{arrows,%
                petri,%
                topaths,
                calc,
                positioning,
                shapes.geometric}%

%
\definecolor{bblue}{HTML}{7f9aff}
\definecolor{rred}{HTML}{870000}
\definecolor{ggreen}{HTML}{9BBB59}
\definecolor{ppurple}{HTML}{9F4C7C}

\pgfplotsset{compat=1.11,
    /pgfplots/ybar legend/.style={
    /pgfplots/legend image code/.code={%
       \draw[##1,/tikz/.cd,yshift=-0.25em]
        (0cm,0cm) rectangle (3pt,0.8em);},
   },
}

\begin{tikzpicture}[scale=0.8]
    \begin{axis}[
        width  = 0.5*\textwidth,
        height = 8cm,
        major x tick style = transparent,
        ybar = 2*\pgflinewidth,
        bar width=14pt,
        ymajorgrids = true,
        axis x line*=bottom,
        symbolic x coords={Trial 1,Trial 2,Trial 3},
        xtick = data,
        scaled y ticks = false,
        enlarge x limits=0.25,
        ymin=0,
        legend cell align=left,
        legend columns=3,
        legend style={
                at={(0.7,-0.2)},
                anchor=south east,
                column sep=1ex,
                draw=none
        }
    ]

    \addplot[style={bblue,fill=bblue,mark=none},error bars/.cd, y dir=both, y explicit, error bar style=black]
        coordinates {   (Trial 1, 21.022)      += (0,2.396) -= (0,2.396)
                        (Trial 2, 14.731)      += (0,1.303) -= (0,1.303)
                        (Trial 3, 13.346)      += (0,0.915) -= (0,0.915)
                    };
    \addplot[style={ggreen,fill=ggreen,mark=none},error bars/.cd, y dir=both, y explicit, error bar style=black]
        coordinates {   (Trial 1, 20.647)     += (0,2.470) -= (0,2.470)
                        (Trial 2, 16.034)     += (0,1.199) -= (0,1.199)
                        (Trial 3, 14.378)     += (0,0.672) -= (0,0.672)
                    };
    \addplot[style={rred,fill=rred,mark=none},error bars/.cd, y dir=both, y explicit, error bar style=black]
        coordinates {   (Trial 1, 20.171)     += (0,2.690) -= (0,2.690)
                        (Trial 2, 19.757)     += (0,1.669) -= (0,1.669)
                        (Trial 3, 15.342)     += (0,1.919) -= (0,1.919)
                    };

    \legend{MO, AR, MR}
    \end{axis}
    
    \draw (0.45, 5.8) -- (0.45, 6) -- (1.96, 6) -- (1.96, 5.8);
    \draw (2.9, 5.8) -- (2.9, 6) -- (4.41, 6) -- (4.41, 5.8);
    \draw (5.3, 5.8) -- (5.3, 6) -- (6.81, 6) -- (6.81, 5.8);
    
    \draw (1.205, 6) -- (1.205, 6.6) --node[above, yshift=-2.8] {$*$} (6.31, 6.6) -- (6.31, 6);
    \draw (3.655, 6) -- (3.655, 6.3) --node[above, yshift=-2.8] {$*$} (5.81, 6.3) -- (5.81, 6);
    
\end{tikzpicture}

%% file: chart_subj.tex
\usetikzlibrary{arrows,%
                petri,%
                topaths,
                calc,
                positioning,
                shapes.geometric}%

%
\definecolor{bblue}{HTML}{7f9aff}
\definecolor{rred}{HTML}{870000}
\definecolor{ggreen}{HTML}{9BBB59}
\definecolor{ppurple}{HTML}{9F4C7C}

\pgfplotsset{compat=1.11,
    /pgfplots/ybar legend/.style={
    /pgfplots/legend image code/.code={%
       \draw[##1,/tikz/.cd,yshift=-0.25em]
        (0cm,0cm) rectangle (3pt,0.8em);},
   },
}

\begin{tikzpicture}[scale=0.8]
    \begin{axis}[
        width  = 0.5*\textwidth,
        height = 8cm,
        major x tick style = transparent,
        ybar = 2*\pgflinewidth,
        bar width=14pt,
        ymajorgrids = true,
        axis x line*=bottom,
        symbolic x coords={Engagement,Observability,Display interference},
        xtick = data,
        scaled y ticks = false,
        enlarge x limits=0.25,
        ymin=1,
        ymax=7,
        legend cell align=left,
        legend columns=3,
        legend style={
                at={(0.7,-0.2)},
                anchor=south east,
                column sep=1ex,
                draw=none
        }
    ]
    \addplot[style={bblue,fill=bblue,mark=none},error bars/.cd, y dir=both, y explicit, error bar style=black]
        coordinates {   (Engagement,            4.250)      += (0,0.265) -= (0,0.265)
                        (Observability,         6.069)      += (0,0.185) -= (0,0.185)
                        (Display interference,  3.448)      += (0,0.399) -= (0,0.399)
                    };
    \addplot[style={ggreen,fill=ggreen,mark=none},error bars/.cd, y dir=both, y explicit, error bar style=black]
        coordinates {   (Engagement,            4.250)      += (0,0.260) -= (0,0.260)
                        (Observability,         6.483)      += (0,0.137) -= (0,0.137)
                        (Display interference,  1.862)      += (0,0.288) -= (0,0.288)
                    };
    \addplot[style={rred,fill=rred,mark=none},error bars/.cd, y dir=both, y explicit, error bar style=black]
        coordinates {   (Engagement,            4.893)      += (0,0.318) -= (0,0.318)
                        (Observability,         5.621)      += (0,0.240) -= (0,0.240)
                        (Display interference,  4.034)      += (0,0.356) -= (0,0.356)
                    };

    \legend{MO, AR, MR}
    \end{axis}
    
    \draw (0.70, 3.75) -- (0.70, 6.6) --node[above, yshift=-2.8] {$*$} (1.74, 6.6) -- (1.74, 4.5);
    \draw (1.23, 3.75) -- (1.23, 6.3) --node[above, yshift=-2.8] {$*$} (1.74, 6.3) -- (1.74, 4.5);
    
    \draw (3.13, 5.63) -- (3.13, 6.6) --node[above, yshift=-2.8] {$*$} (3.65, 6.6) -- (3.65, 6.0);
    \draw (4.17, 5.20) -- (4.17, 6.3) --node[above, yshift=-2.8] {$*$} (3.65, 6.3);
    
    \draw (5.58, 3.05) -- (5.58, 6.6) --node[above, yshift=-2.8] {$*$} (6.62, 6.6) -- (6.62, 3.63);
    \draw (6.1, 1.23) -- (6.1, 6.3) --node[above, yshift=-2.8] {$*$} (6.62, 6.3) -- (6.62, 3.63);
    
\end{tikzpicture}

%% file: yumi_hri2018_header.bbl

\begin{thebibliography}{26}


\ifx \showCODEN    \undefined \def \showCODEN     #1{\unskip}     \fi
\ifx \showDOI      \undefined \def \showDOI       #1{#1}\fi
\ifx \showISBNx    \undefined \def \showISBNx     #1{\unskip}     \fi
\ifx \showISBNxiii \undefined \def \showISBNxiii  #1{\unskip}     \fi
\ifx \showISSN     \undefined \def \showISSN      #1{\unskip}     \fi
\ifx \showLCCN     \undefined \def \showLCCN      #1{\unskip}     \fi
\ifx \shownote     \undefined \def \shownote      #1{#1}          \fi
\ifx \showarticletitle \undefined \def \showarticletitle #1{#1}   \fi
\ifx \showURL      \undefined \def \showURL       {\relax}        \fi
\providecommand\bibfield[2]{#2}
\providecommand\bibinfo[2]{#2}
\providecommand\natexlab[1]{#1}
\providecommand\showeprint[2][]{arXiv:#2}

\bibitem[\protect\citeauthoryear{Admoni, Weng, and Scassellati}{Admoni
  et~al\mbox{.}}{2016}]%
        {admoni2016modeling}
\bibfield{author}{\bibinfo{person}{Henny Admoni}, \bibinfo{person}{Thomas
  Weng}, {and} \bibinfo{person}{Brian Scassellati}.}
  \bibinfo{year}{2016}\natexlab{}.
\newblock \showarticletitle{Modeling communicative behaviors for object
  references in human-robot interaction}. In \bibinfo{booktitle}{{\em Robotics
  and Automation (ICRA), 2016 IEEE International Conference on}}. IEEE,
  \bibinfo{pages}{3352--3359}.
\newblock


\bibitem[\protect\citeauthoryear{Andersen, Madsen, Moeslund, and Amor}{Andersen
  et~al\mbox{.}}{2016}]%
        {andersen2016}
\bibfield{author}{\bibinfo{person}{Rasmus~S Andersen}, \bibinfo{person}{Ole
  Madsen}, \bibinfo{person}{Thomas~B Moeslund}, {and} \bibinfo{person}{Heni~Ben
  Amor}.} \bibinfo{year}{2016}\natexlab{}.
\newblock \showarticletitle{Projecting robot intentions into human
  environments}. In \bibinfo{booktitle}{{\em Robot and Human Interactive
  Communication (RO-MAN), 2016 25th IEEE International Symposium on}}. IEEE,
  \bibinfo{pages}{294--301}.
\newblock


\bibitem[\protect\citeauthoryear{Bolt}{Bolt}{1980}]%
        {bolt1980put}
\bibfield{author}{\bibinfo{person}{Richard~A Bolt}.}
  \bibinfo{year}{1980}\natexlab{}.
\newblock \bibinfo{booktitle}{{\em "Put-that-there": Voice and gesture at the
  graphics interface}}. Vol.~\bibinfo{volume}{14}.
\newblock \bibinfo{publisher}{ACM}.
\newblock


\bibitem[\protect\citeauthoryear{Chadalavada, Andreasson, Krug, and
  Lilienthal}{Chadalavada et~al\mbox{.}}{2015}]%
        {chadalavada2015}
\bibfield{author}{\bibinfo{person}{Ravi~Teja Chadalavada},
  \bibinfo{person}{Henrik Andreasson}, \bibinfo{person}{Robert Krug}, {and}
  \bibinfo{person}{Achim~J Lilienthal}.} \bibinfo{year}{2015}\natexlab{}.
\newblock \showarticletitle{That's on my mind! robot to human intention
  communication through on-board projection on shared floor space}. In
  \bibinfo{booktitle}{{\em Mobile Robots (ECMR), 2015 European Conference on}}.
  IEEE, \bibinfo{pages}{1--6}.
\newblock


\bibitem[\protect\citeauthoryear{Chai, She, Fang, Ottarson, Littley, Liu, and
  Hanson}{Chai et~al\mbox{.}}{2014}]%
        {chai2014collaborative}
\bibfield{author}{\bibinfo{person}{Joyce~Y Chai}, \bibinfo{person}{Lanbo She},
  \bibinfo{person}{Rui Fang}, \bibinfo{person}{Spencer Ottarson},
  \bibinfo{person}{Cody Littley}, \bibinfo{person}{Changsong Liu}, {and}
  \bibinfo{person}{Kenneth Hanson}.} \bibinfo{year}{2014}\natexlab{}.
\newblock \showarticletitle{Collaborative effort towards common ground in
  situated human-robot dialogue}. In \bibinfo{booktitle}{{\em Proceedings of
  the 2014 ACM/IEEE international conference on Human-robot interaction}}. ACM,
  \bibinfo{pages}{33--40}.
\newblock


\bibitem[\protect\citeauthoryear{Chitta, Sucan, and Cousins}{Chitta
  et~al\mbox{.}}{2012}]%
        {chitta2012moveit}
\bibfield{author}{\bibinfo{person}{Sachin Chitta}, \bibinfo{person}{Ioan
  Sucan}, {and} \bibinfo{person}{Steve Cousins}.}
  \bibinfo{year}{2012}\natexlab{}.
\newblock \showarticletitle{Moveit![ROS topics]}.
\newblock \bibinfo{journal}{{\em IEEE Robotics \& Automation Magazine\/}}
  \bibinfo{volume}{19}, \bibinfo{number}{1} (\bibinfo{year}{2012}),
  \bibinfo{pages}{18--19}.
\newblock


\bibitem[\protect\citeauthoryear{Clark}{Clark}{1996}]%
        {clark1996using}
\bibfield{author}{\bibinfo{person}{Herbert~H Clark}.}
  \bibinfo{year}{1996}\natexlab{}.
\newblock \bibinfo{booktitle}{{\em Using language}}.
\newblock \bibinfo{publisher}{Cambridge university press}.
\newblock


\bibitem[\protect\citeauthoryear{Frank, Moorhead, and Kapila}{Frank
  et~al\mbox{.}}{2017}]%
        {frank2016}
\bibfield{author}{\bibinfo{person}{Jared~A. Frank}, \bibinfo{person}{Matthew
  Moorhead}, {and} \bibinfo{person}{Vikram Kapila}.}
  \bibinfo{year}{2017}\natexlab{}.
\newblock \showarticletitle{Mobile Mixed-Reality Interfaces That Enhance
  Human-Robot Interaction in Shared Spaces}.
\newblock \bibinfo{journal}{{\em Frontiers in Robotics and AI\/}}
  \bibinfo{volume}{4} (\bibinfo{year}{2017}), \bibinfo{pages}{20}.
\newblock
\showISSN{2296-9144}
\showDOI{%
\url{https://doi.org/10.3389/frobt.2017.00020}}


\bibitem[\protect\citeauthoryear{Guadarrama, Riano, Golland, Go, Jia, Klein,
  Abbeel, Darrell, et~al\mbox{.}}{Guadarrama et~al\mbox{.}}{2013}]%
        {guadarrama2013grounding}
\bibfield{author}{\bibinfo{person}{Sergio Guadarrama}, \bibinfo{person}{Lorenzo
  Riano}, \bibinfo{person}{Dave Golland}, \bibinfo{person}{Daniel Go},
  \bibinfo{person}{Yangqing Jia}, \bibinfo{person}{Dan Klein},
  \bibinfo{person}{Pieter Abbeel}, \bibinfo{person}{Trevor Darrell},
  {et~al\mbox{.}}} \bibinfo{year}{2013}\natexlab{}.
\newblock \showarticletitle{Grounding spatial relations for human-robot
  interaction}. In \bibinfo{booktitle}{{\em Intelligent Robots and Systems
  (IROS), 2013 IEEE/RSJ International Conference on}}. IEEE,
  \bibinfo{pages}{1640--1647}.
\newblock


\bibitem[\protect\citeauthoryear{Hough and Schlangen}{Hough and
  Schlangen}{2017}]%
        {hough2017s}
\bibfield{author}{\bibinfo{person}{Julian Hough} {and} \bibinfo{person}{David
  Schlangen}.} \bibinfo{year}{2017}\natexlab{}.
\newblock \showarticletitle{It's Not What You Do, It's How You Do It: Grounding
  Uncertainty for a Simple Robot}. In \bibinfo{booktitle}{{\em Proceedings of
  the 2017 ACM/IEEE International Conference on Human-Robot Interaction}}. ACM,
  \bibinfo{pages}{274--282}.
\newblock


\bibitem[\protect\citeauthoryear{Kennington and Schlangen}{Kennington and
  Schlangen}{2017}]%
        {kennington2017simple}
\bibfield{author}{\bibinfo{person}{Casey Kennington} {and}
  \bibinfo{person}{David Schlangen}.} \bibinfo{year}{2017}\natexlab{}.
\newblock \showarticletitle{A simple generative model of incremental reference
  resolution for situated dialogue}.
\newblock \bibinfo{journal}{{\em Computer Speech \& Language\/}}
  \bibinfo{volume}{41} (\bibinfo{year}{2017}), \bibinfo{pages}{43--67}.
\newblock


\bibitem[\protect\citeauthoryear{Li, Scalise, Admoni, Rosenthal, and
  Srinivasa}{Li et~al\mbox{.}}{2016}]%
        {li2016spatial}
\bibfield{author}{\bibinfo{person}{Shen Li}, \bibinfo{person}{Rosario Scalise},
  \bibinfo{person}{Henny Admoni}, \bibinfo{person}{Stephanie Rosenthal}, {and}
  \bibinfo{person}{Siddhartha~S Srinivasa}.} \bibinfo{year}{2016}\natexlab{}.
\newblock \showarticletitle{Spatial references and perspective in natural
  language instructions for collaborative manipulation}. In
  \bibinfo{booktitle}{{\em Robot and Human Interactive Communication (RO-MAN),
  2016 25th IEEE International Symposium on}}. IEEE, \bibinfo{pages}{44--51}.
\newblock


\bibitem[\protect\citeauthoryear{Lombard, Ditton, and Weinstein}{Lombard
  et~al\mbox{.}}{2009}]%
        {lombard2009}
\bibfield{author}{\bibinfo{person}{Matthew Lombard}, \bibinfo{person}{Theresa~B
  Ditton}, {and} \bibinfo{person}{Lisa Weinstein}.}
  \bibinfo{year}{2009}\natexlab{}.
\newblock \showarticletitle{Measuring presence: the temple presence inventory}.
  In \bibinfo{booktitle}{{\em Proceedings of the 12th Annual International
  Workshop on Presence}}. \bibinfo{pages}{1--15}.
\newblock


\bibitem[\protect\citeauthoryear{Mehlmann, H{\"a}ring, Janowski, Baur, Gebhard,
  and Andr{\'e}}{Mehlmann et~al\mbox{.}}{2014}]%
        {mehlmann2014exploring}
\bibfield{author}{\bibinfo{person}{Gregor Mehlmann}, \bibinfo{person}{Markus
  H{\"a}ring}, \bibinfo{person}{Kathrin Janowski}, \bibinfo{person}{Tobias
  Baur}, \bibinfo{person}{Patrick Gebhard}, {and} \bibinfo{person}{Elisabeth
  Andr{\'e}}.} \bibinfo{year}{2014}\natexlab{}.
\newblock \showarticletitle{Exploring a model of gaze for grounding in
  multimodal HRI}. In \bibinfo{booktitle}{{\em Proceedings of the 16th
  International Conference on Multimodal Interaction}}. ACM,
  \bibinfo{pages}{247--254}.
\newblock


\bibitem[\protect\citeauthoryear{Paul, Arkin, Roy, and Howard}{Paul
  et~al\mbox{.}}{[n. d.]}]%
        {paulgrounding}
\bibfield{author}{\bibinfo{person}{Rohan Paul}, \bibinfo{person}{Jacob Arkin},
  \bibinfo{person}{Nicholas Roy}, {and} \bibinfo{person}{Thomas Howard}.}
  \bibinfo{year}{[n. d.]}\natexlab{}.
\newblock \showarticletitle{Grounding Abstract Spatial Concepts for Language
  Interaction with Robots}.
\newblock  (\bibinfo{year}{[n. d.]}).
\newblock


\bibitem[\protect\citeauthoryear{Pereira, Carter, Leite, Mars, and
  Lehman}{Pereira et~al\mbox{.}}{2017}]%
        {pereira2017}
\bibfield{author}{\bibinfo{person}{Ande Pereira}, \bibinfo{person}{Elizabeth~J.
  Carter}, \bibinfo{person}{Iolanda Leite}, \bibinfo{person}{John Mars}, {and}
  \bibinfo{person}{Jill~Fain Lehman}.} \bibinfo{year}{2017}\natexlab{}.
\newblock \showarticletitle{Augmented Reality Dialog Interface for Multimodal
  Teleoperation}. In \bibinfo{booktitle}{{\em Robot and Human Interactive
  Communication (RO-MAN), 2017 26th IEEE International Symposium on Robot and
  Human Interactive Communication}}. IEEE.
\newblock


\bibitem[\protect\citeauthoryear{Renner, Pfeiffer, and Wachsmuth}{Renner
  et~al\mbox{.}}{2014}]%
        {renner2014spatial}
\bibfield{author}{\bibinfo{person}{Patrick Renner}, \bibinfo{person}{Thies
  Pfeiffer}, {and} \bibinfo{person}{Ipke Wachsmuth}.}
  \bibinfo{year}{2014}\natexlab{}.
\newblock \showarticletitle{Spatial references with gaze and pointing in shared
  space of humans and robots}. In \bibinfo{booktitle}{{\em International
  Conference on Spatial Cognition}}. Springer, \bibinfo{pages}{121--136}.
\newblock


\bibitem[\protect\citeauthoryear{Rohan~Paul}{Rohan~Paul}{2017}]%
        {ijcai2017-629}
\bibfield{author}{\bibinfo{person}{Sue Felshin Boris Katz Nicholas~Roy
  Rohan~Paul, Andrei~Barbu}.} \bibinfo{year}{2017}\natexlab{}.
\newblock \showarticletitle{Temporal Grounding Graphs for Language
  Understanding with Accrued Visual-Linguistic Context}. In
  \bibinfo{booktitle}{{\em Proceedings of the Twenty-Sixth International Joint
  Conference on Artificial Intelligence, {IJCAI-17}}}.
  \bibinfo{pages}{4506--4514}.
\newblock
\showDOI{%
\url{https://doi.org/10.24963/ijcai.2017/629}}


\bibitem[\protect\citeauthoryear{Rosen, Whitney, Phillips, Chien, Tompkin,
  Konidaris, and Tellex}{Rosen et~al\mbox{.}}{2017}]%
        {rosen2017}
\bibfield{author}{\bibinfo{person}{Eric Rosen}, \bibinfo{person}{David
  Whitney}, \bibinfo{person}{Elizabeth Phillips}, \bibinfo{person}{Gary Chien},
  \bibinfo{person}{James Tompkin}, \bibinfo{person}{George Konidaris}, {and}
  \bibinfo{person}{Stefanie Tellex}.} \bibinfo{year}{2017}\natexlab{}.
\newblock \showarticletitle{Communicating Robot Arm Motion Intent Through Mixed
  Reality Head-mounted Displays}.
\newblock \bibinfo{journal}{{\em arXiv preprint arXiv:1708.03655\/}}
  (\bibinfo{year}{2017}).
\newblock


\bibitem[\protect\citeauthoryear{Ruffaldi, Brizzi, Tecchia, and
  Bacinelli}{Ruffaldi et~al\mbox{.}}{2016}]%
        {Ruffaldi2016}
\bibfield{author}{\bibinfo{person}{Emanuele Ruffaldi}, \bibinfo{person}{Filippo
  Brizzi}, \bibinfo{person}{Franco Tecchia}, {and} \bibinfo{person}{Sandro
  Bacinelli}.} \bibinfo{year}{2016}\natexlab{}.
\newblock \bibinfo{booktitle}{{\em Third Point of View Augmented Reality for
  Robot Intentions Visualization}}.
\newblock \bibinfo{publisher}{Springer International Publishing},
  \bibinfo{address}{Cham}, \bibinfo{pages}{471--478}.
\newblock
\showDOI{%
\url{https://doi.org/10.1007/978-3-319-40621-3_35}}


\bibitem[\protect\citeauthoryear{Saupp{\'e} and Mutlu}{Saupp{\'e} and
  Mutlu}{2014}]%
        {sauppe2014robot}
\bibfield{author}{\bibinfo{person}{Allison Saupp{\'e}} {and}
  \bibinfo{person}{Bilge Mutlu}.} \bibinfo{year}{2014}\natexlab{}.
\newblock \showarticletitle{Robot deictics: How gesture and context shape
  referential communication}. In \bibinfo{booktitle}{{\em Proceedings of the
  2014 ACM/IEEE international conference on Human-robot interaction}}. ACM,
  \bibinfo{pages}{342--349}.
\newblock


\bibitem[\protect\citeauthoryear{Skantze}{Skantze}{2010}]%
        {skantze2010jindigo}
\bibfield{author}{\bibinfo{person}{Gabriel Skantze}.}
  \bibinfo{year}{2010}\natexlab{}.
\newblock \showarticletitle{Jindigo: a Java-based framework for incremental
  dialogue systems}.
\newblock \bibinfo{journal}{{\em Proceedings of Interspeech. submitted, www.
  jidingo. net\/}} (\bibinfo{year}{2010}).
\newblock


\bibitem[\protect\citeauthoryear{Steels and Hild}{Steels and Hild}{2012}]%
        {steels2012language}
\bibfield{author}{\bibinfo{person}{Luc Steels} {and} \bibinfo{person}{Manfred
  Hild}.} \bibinfo{year}{2012}\natexlab{}.
\newblock \bibinfo{booktitle}{{\em Language grounding in robots}}.
\newblock \bibinfo{publisher}{Springer Science \& Business Media}.
\newblock


\bibitem[\protect\citeauthoryear{Whitney, Rosen, MacGlashan, Wong, and
  Tellex}{Whitney et~al\mbox{.}}{2017}]%
        {whitney2017}
\bibfield{author}{\bibinfo{person}{David Whitney}, \bibinfo{person}{Eric
  Rosen}, \bibinfo{person}{James MacGlashan}, \bibinfo{person}{Lawson~LS Wong},
  {and} \bibinfo{person}{Stefanie Tellex}.} \bibinfo{year}{2017}\natexlab{}.
\newblock \showarticletitle{Reducing errors in object-fetching interactions
  through social feedback}. In \bibinfo{booktitle}{{\em 2017 IEEE International
  Conference on Robotics and Automation (ICRA)}}. IEEE,
  \bibinfo{pages}{1006--1013}.
\newblock


\bibitem[\protect\citeauthoryear{Winograd}{Winograd}{1972}]%
        {winograd1972understanding}
\bibfield{author}{\bibinfo{person}{Terry Winograd}.}
  \bibinfo{year}{1972}\natexlab{}.
\newblock \showarticletitle{Understanding natural language}.
\newblock \bibinfo{journal}{{\em Cognitive psychology\/}} \bibinfo{volume}{3},
  \bibinfo{number}{1} (\bibinfo{year}{1972}), \bibinfo{pages}{1--191}.
\newblock


\bibitem[\protect\citeauthoryear{Witmer and Singer}{Witmer and Singer}{1998}]%
        {witmer1998}
\bibfield{author}{\bibinfo{person}{Bob~G Witmer} {and}
  \bibinfo{person}{Michael~J Singer}.} \bibinfo{year}{1998}\natexlab{}.
\newblock \showarticletitle{Measuring presence in virtual environments: A
  presence questionnaire}.
\newblock \bibinfo{journal}{{\em Presence: Teleoperators and virtual
  environments\/}} \bibinfo{volume}{7}, \bibinfo{number}{3}
  (\bibinfo{year}{1998}), \bibinfo{pages}{225--240}.
\newblock


\end{thebibliography}
